\documentclass[conference]{IEEEtran}
\IEEEoverridecommandlockouts
\usepackage{cite}
\usepackage{amsmath,amssymb,amsfonts}
\usepackage{algorithmic}
\usepackage{graphicx}
\usepackage{textcomp}
\usepackage{xcolor}
\usepackage{subcaption}

\usepackage[skip=2pt]{caption}
\begin{document}

\title{%
Teacher-Student Representational Alignment for Reinforcement Learning-Driven Imitation Learning 
}

\author{\IEEEauthorblockN{Meraj Mammadov}
\IEEEauthorblockA{
\textit{Department of Computer Science} \\
\textit{Örebro University}\\
meraj.mammadov@oru.se}
\and
\IEEEauthorblockN{Pedro Zuidberg Dos Martires}
\IEEEauthorblockA{
\textit{Department of Computer Science} \\
\textit{Örebro University}\\
pedro.zuidberg-dos-martires@oru.se}
\and
\IEEEauthorblockN{Johannes Andreas Stork}
\IEEEauthorblockA{
\textit{Department of Computer Science} \\
\textit{Örebro University}\\
johannesandreas.stork@oru.se}
}

\maketitle
\thispagestyle{empty}
\pagestyle{plain}

\begin{abstract}
Imitation learning (IL) from a state-based reinforcement learning (RL) policy is a common approach to overcome the curse of dimensionality in complex and high-dimensional observation spaces prevalent in robotics. This paper addresses the irreducible imitation gap that emerges when teacher and student are learned in isolation, and the teacher policy has the liberty to rely on privileged state information that the student cannot infer from its observations. Instead of improving poor student performance with RL finetuning after IL, which often requires a whole new training setup, we propose a novel algorithm which learns a shared embedding space that hides agent-specific observations and thus trains imitable teacher policies by construction. We train the shared embedding space with self-supervised contrastive learning in parallel to the teacher policy and prevent it from extracting private information by limiting its gradients from updating the encoder networks. We perform evaluations on several example domains and compare to state-of-the-art baselines showing that our algorithm enables higher student performance with substantially reduced imitation gap.
\end{abstract}


\section{Introduction}
\label{sec:introduction}

\begin{figure}
\centering
\includegraphics[width=0.95 \linewidth]{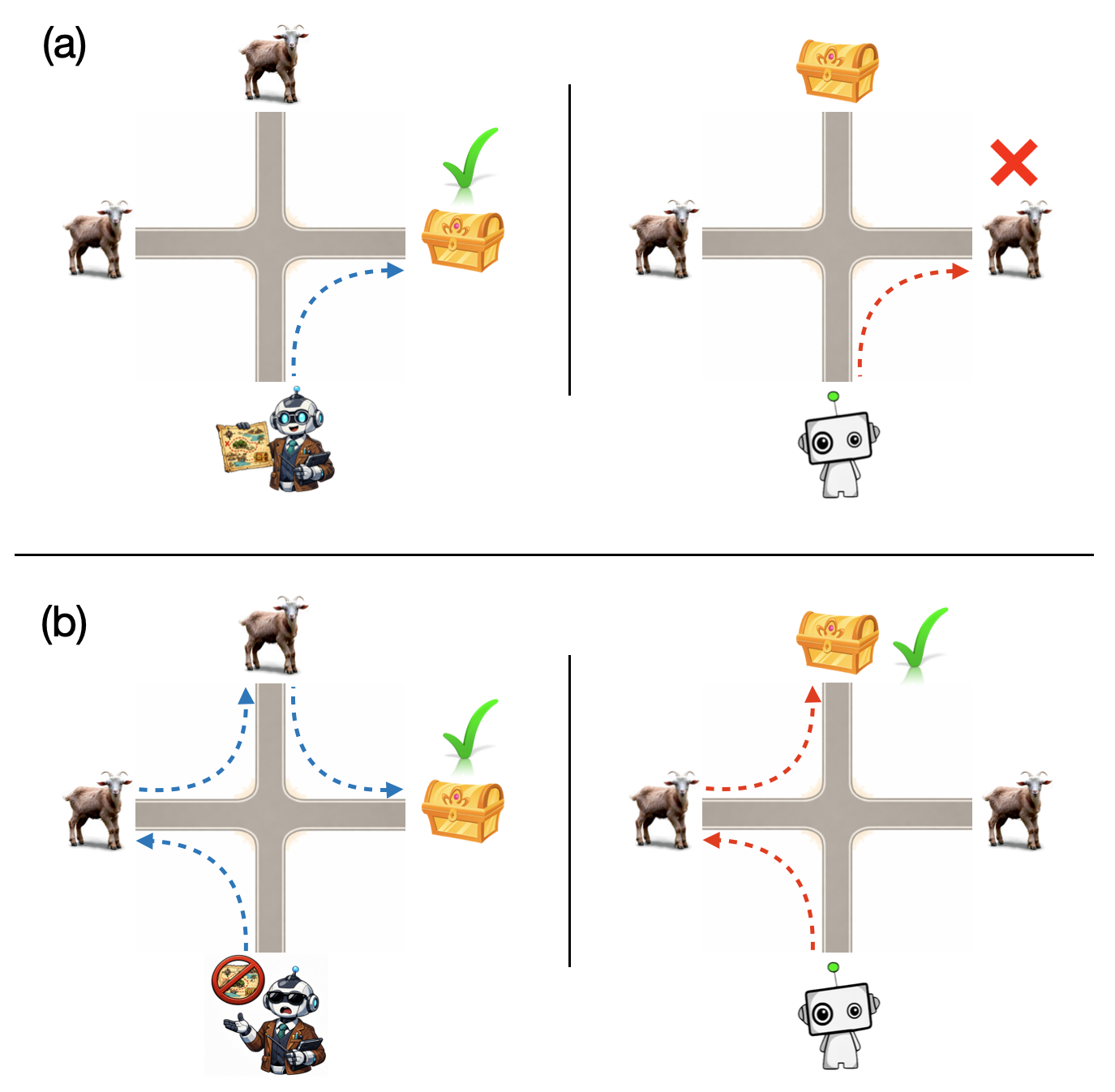}
\caption{%
Illustration of the imitation gap: (a) Training the teacher (left) and student (right) in isolation leads to non-imitable teacher behavior in presence of private information, such as the goal location. 
(b) Our approach automatically hides the private information from the teacher, making the learned teacher behavior imitable by the student.
}
\label{fig:illustration}
\end{figure}

Learning behavior policies in high-dimensional observation spaces using reinforcement learning (RL) suffers from high sample complexity \cite{yarats2022mastering}, while more efficient approaches such as imitation learning (IL) require access to costly expert demonstrations. Recent work shows the effectiveness of first training an RL-based teacher on low-dimensional analytical states and then using IL to distill the behavior to a student policy operating on high-dimensional observation space \cite{chen2019learning}. For this to be efficient, the teacher is often given access to highly informative and task-relevant state variables and dense reward signals. While providing more information to the teacher agent leads to higher teacher performance and faster convergence, it also enables the teacher to rely on information that is private to it and cannot be derived from student observations (See Fig. \ref{fig:illustration} for an illustration). This can lead to a non-reducible imitation gap \cite{walsman2023impossibly} in student performance during distillation, which is often addressed by an additional RL finetuning phase on the student policy \cite{weihs2021bridging, schmitt2018kickstarting}.

As the additional RL finetuning step typically involves the original difficulties with RL training, it is more desirable to train the teacher policy in a way that the student can imitate it from the beginning. 
To this end, we propose a combined RL and IL learning approach that automatically hides the teacher-specific private information from the teacher's observations and enables the student to imitate it by construction.
Unlike previous work that constrains the teacher's action space \cite{monaci2022dipcan} or modify its already sensitive reward signal \cite{song2023learning, messikommer2025student}, we pose a representation learning problem: Our goal is to learn a low-dimensional common representation space that only retains the shared part of the teacher and student observations while removing agent-specific state variables.
The learned joint representation is used by the teacher during its online RL training on the task and by the student during the imitation phase. The embedding space and the policy are trained in alternating phases with separate objectives to prevent the teacher from extracting observations specific to its observation space. We apply additional alignment and stability losses to the embedding space to encourage representational similarity between aligned observations and enable stable policy learning.

The advantage of our approach is that the teacher learns a policy that can be directly imitated by the student without changing the teacher's reward structure, which can break  task learning or lead to reward hacking by the agent. The change from teacher observations to joint representation requires the teacher to maximize the original reward function by exploiting only the information that is shared between teacher and student which generates behaviors that the student can imitate. Furthermore, our choice of positive and negative samples for the contrastive learning objective preserves small variations in the consecutive states and enables the agent to learn more refined policies.
Our evaluation on a series of challenging environments shows that our approach can  distill successful student policies. Comparisons to state-of-the-art baseline approaches show that our approach compares favorably in terms of student performance. An ablation study shows the benefits of the different loss terms.

Our contributions can be summarized as follows:

\begin{itemize}
    \item We propose a novel, task-agnostic method to bridge the imitation gap between RL-based teacher and IL-based student policies.

    \item The proposed method is a plug-and-play adaptation to existing frameworks with no modification to the reward function and minimal hyperparameter tuning.

    \item We evaluate our approach in two environments designed to expose the imitation gap and show that it consistently outperforms strong baseline methods.
\end{itemize}

\section{Related Work}
\label{sec:relatedwork}

Imitation learning has been widely adopted in robotics to overcome the curse of dimensionality and accelerate policy learning. Works such as \cite{tordesillas2023deep} cut computational costs by mimicking demanding planning algorithms using neural networks while \cite{zhao2025aloha} imitates human experts to learn fine-grained manipulation tasks. Despite their success, the traditional IL algorithms rely on access to an interactive expert algorithm or the existence of collected expert demonstrations. A recent paradigm in IL called "learning by cheating" \cite{chen2019learning} overcomes this limitation by obtaining an interactive teacher policy by applying RL on low-dimensional analytical states, which is later combined with IL methods to learn high-dimensional policies.

An irreducible performance gap arises between the teacher and student agents when the teacher policy makes its decisions based on state variables that are unobservable to the student. \cite{xing2024bootstrapping} recovers the student's performance by additional RL finetuning on the distilled student policy, however bootstrapping RL policies without a coupled critic network is shown to be ineffective and possibly lead to inferior performance \cite{nakamoto2023calql}. \cite{weihs2021bridging} instead regards the teacher signals as a guidance during the student's own RL training and combines it with the original task objective. \cite{walsman2023impossibly} trains a value function for the distilled suboptimal policy and uses it to guide the exploration of a student policy trained from scratch.

Instead, our work aims at training the teacher policies in an imitable way from construction. \cite{song2023learning} designs dense reward signals by exploiting the environment knowledge to motivate imitable teacher behaviors. However, such knowledge is generally not available and require heavy feature engineering to achieve intended policies. A work closely related to our setup is SITT \cite{messikommer2025student}, which trains the teacher policy in parallel with the student policy to avoid learning behaviors that the student cannot imitate. It requires no prior knowledge of the environment and avoids post-hoc fine-tuning of the student, but modifies the original reward function and introduces a trade-off coefficient that is difficult to tune and requires careful balancing. Our method avoids feature engineering and modifications to the reward function by constructing a common representation space with contrastive learning for both policies to operate on. 

Similar ideas in contrastive learning has been previously used in imitation learning settings to enable policy transfer under various distributional shifts.  \cite{giammarino2025visually} uses contrastive learning to construct an embedding space that is invariant to different augmentations of the input space caused by changing background and lighting conditions. An adversarial IL policy trained on this latent space shows substantially reduced imitation gap compared to direct baselines. Similarly, \cite{xing2024contrastive} learns an embedding space under which the real and simulated versions of an environment scene look similar, which is later used to transfer skills learned in one domain to another. Their method relies on the domain knowledge to select the contrastive pairs and uses a heuristic-based similarity measure to construct the embedding space. Our algorithm however is fully task-agnostic and requires no domain knowledge, which enables it to be easily integrated into existing IL frameworks.

\section{Background}
\label{sec:background}

We formalize the RL task a Markov decision process (MDP) characterized by the tuple $(\mathcal O_T, \mathcal A, P, R, \gamma)$. $\mathcal O_T$ denotes the observation spaces of the teacher. $\mathcal A$ is the action space, $P(s^\prime \mid s, a)$ is the transition probability to state $s^\prime$ given action $a$ is taken in state $s$, $R(s, a)$ is the reward function, and finally $\gamma$ is the discount factor. The MDP for the student is identical but uses the observation space $\mathcal O_S$ instead.
Teacher and student policies, $\pi_T$ and $\pi_S$ map their respective observations to a probability distribution over the action space: $a_i \sim \pi_i(\cdot \mid o_i), \;i\in \{T, S\}$. In our setup, $\pi_T$ is trained with reinforcement learning to guide $\pi_S$ using imitation learning.

\subsection{Reinforcement learning}

Reinforcement learning is a framework to learn optimal policies through interaction with an environment. We train the teacher policy $\pi_T$ with RL to maximize the expected discounted return:
\begin{equation}
J(\pi_T) =
\mathbb{E}_{\tau \sim \pi_T, \; a_t \sim \pi_T}
\left[
\sum_{t=0}^{\infty} \gamma^t R(s_t, a_t)
\right],
\end{equation}
where $s_{t+1} \sim P(\cdot \mid s_t, a_t)$, and $\tau = (o_0, a_0, r_0, o_1, a_1, r_1 \dots)$ denotes a trajectory generated by the teacher policy interacting with the environment. We use Proximal Policy Optimization (PPO) \cite{schulman2017proximal} algorithm throughout our experiments.

\subsection{Contrastive learning}
Contrastive learning (CL) is a self-supervised representation learning method to learn rich data representations in the absence of direct supervision signals. CL achieves this by bringing similar samples closer in the embedding space while pushing dissimilar ones apart. Given an anchor sample and a corresponding positive example, along with a set of negative samples, the objective is typically formulated using a similarity-based loss such as the triplet or InfoNCE \cite{chen2020simple} losses. Let $z_i$ and $z_j$ denote latent representations of a positive pair, and $\mathcal{N}_i$ the set of negatives for $i$. The InfoNCE objective is:
\begin{equation}
\mathcal{L}_{\text{InfoNCE}} =
-\log
\frac{\exp(\mathrm{sim}(z_i, z_j)/\tau)}
{\sum_{z_k \in \{z_j\} \cup \mathcal{N}_i} \exp(\mathrm{sim}(z_i, z_k)/\tau)},
\end{equation}
where $\mathrm{sim}(\cdot, \cdot)$ denotes a measure of similarity between the embeddings (e.g., cosine similarity), and $\tau$ is a temperature parameter.

\section{Method}
\label{sec:method}

Our training setup consists of two separate modules with differing objectives. The first module uses contrastive learning to learn a mapping from each observation space to a common latent space that only contains information that is available to both agents while hiding out the agent-specific private information. The second module is trained with reinforcement learning on the learned common embedding space to maximize the task objective.

We iteratively train each module in two phases: (1) In phase one, we learn the embedding space with contrastive learning trained on aligned dataset of observations from both agents, collected by rolling out the teacher policy. (2) In the second phase, the policy network is trained with RL on the latent embeddings collected by the teacher to maximize the task objective. To prevent the policy gradients from extracting private information from the teacher observations, the policy gradients from the second phase are only allowed to update the policy network and are not passed through the embedding networks.

\subsection{Problem setup}
\begin{figure}
\centering
\includegraphics[scale=0.15]{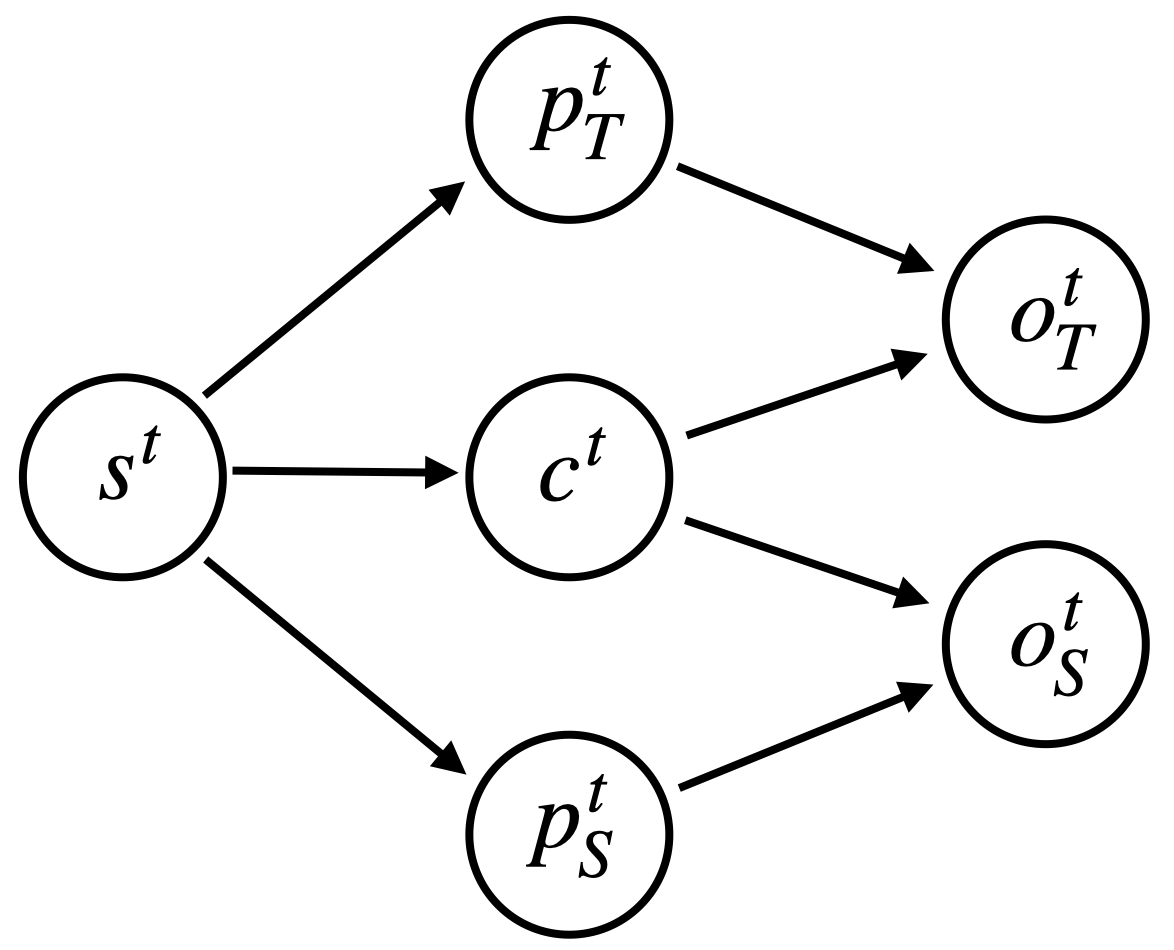}
\caption{We adopt a multi--view approach to the teacher--student setup where the observations from each agent ($o^t_T$ and $o^t_S$) are different views of the same state with their own private ($p^t_T$ and $p^t_S$) and common ($c^t$) components.}
\label{fig:decomposition}
\end{figure}

We adopt a multi-view approach to the imitation learning problem, where the teacher and student agents observe the environment through different modalities. We assume that their paired observations, denoted as $ o_T^t \in \mathcal{O}_T $ and $ o_S^t \in \mathcal{O}_S $, are two distinct views of the same underlying state $ s^t \in \mathcal{S} $:
\begin{equation}
\begin{aligned}
o_T^t = f_T(s^t), \quad & o_S^t = f_S(s^t)
\end{aligned}
\end{equation}
where $f_T$ and $f_S$ are observation functions that map the environment state to respective observations, which are readily available in simulation environments. Since each view captures only a subset of the state's information with a certain overlap, we model each observation as a combination of their common $\left( c^t\right)$ and private components $\left( p_T^t, \; p_S^t\right)$:
\begin{equation}
\begin{aligned}
o_T^t = [c^t, \; p_T^t], \quad & o_S^t = [c^t, \; p_S^t].
\end{aligned}
\label{eq:decomposition}
\end{equation}
The common component $c^t$ contains information that is available to both modalities, e.g. the relative obstacle locations in a collision avoidance task, while the private components $p_T^t$ and $p_S^t$ contain modality-specific details, which are typically not retrievable from the other modality (see Fig. \ref{fig:decomposition} for overview). Following the widely adopted literature in multi-view representation learning \cite{yao2024multiview, lyu2022latent}, we assume that the private components are conditionally independent given the shared variable $c^t$:
\begin{equation}
    p_T^t \perp p_S^t \mid c^t
\label{eq:independence}
\end{equation}
Under these conditions, we aim to retrieve $c^t$ from the original teacher and student observations using contrastive learning.

\subsection{Learning common latent space}
A teacher observation $o_T^t$ at timestep $t$ contains the same common content $c^t$ as the student observation at the same timestep $o_S^t$, which in general is different from the observations collected at any other timestep $t': t' \neq t$. Based on this knowledge, the latent embeddings from two paired observations should reside close to each other while the embeddings for the non-paired observations should be dissimilar. We construct such a latent space using the contrastive self-supervised InfoNCE loss \cite{chen2020simple}, which has been shown to recover the underlying shared signals up to an invertible transformation \cite{vonkugelgen2021contentstyle} under our assumptions (\ref{eq:decomposition}) and (\ref{eq:independence}):
\begin{equation}
\begin{aligned}
\mathcal{L}_{\mathrm{contrastive}}
&=
-\frac{1}{2N}\sum_{i=1}^{N}
\Bigg[
\log
\frac{\exp((z_i^T)^\top z_i^S/\tau)}
{\sum_{j=1}^{N}\exp((z_i^T)^\top z_j^S/\tau)}
\\
&\hspace{+1.2em}+
\log
\frac{\exp((z_i^S)^\top z_i^T/\tau)}
{\sum_{j=1}^{N}\exp((z_i^S)^\top z_j^T/\tau)}
\Bigg]
\end{aligned}
\label{eq:contrastive}
\end{equation}
where the unit-norm embeddings $z_S^i$ and $z_T^i$ are processed by separate encoder networks and their similarity is measured by a dot product: $\mathrm{sim}(z_T^i, z_S^j)=(z_T^i)^\top z_S^j.$ $\tau$ is a learnable parameter that controls the sharpness of the contrastive distribution. Note that unlike time-contrastive representation learning methods that consider all the observations from a fixed time window as positives, \cite{xing2024contrastive}, our objective considers only the aligned student observation as a positive while any other observation in the training batch is pushed away. This distinction is critical for our method as it enables the policy network to distinguish subtle variations among consecutive states and generate distinct actions accordingly.

\subsection{Aligning learned embeddings}
The same policy network in our setup operates on the latent embeddings of the teacher observations during training and the student observations during testing. Since RL policies are known to be sensitive to small changes in the embedding space \cite{cobbe2019quantifying}, we require the embeddings coming from a paired set of observations to be similar to each other. While this similarity ($\mathrm{sim}(z_T^i, z_S^i)$) emerges implicitly at the optimum of $\mathcal L_\text{contrastive}$, we introduce an explicit alignment loss to directly maximize it and guide the training of the embedding space:
\begin{equation}
    \mathcal L_\text{alignment} = -\frac{1}{N} \sum_{i=1}^{N} \mathrm{sim}(z_T^i, z_S^i),
\end{equation}
where the similarity is measured by a dot product between the embeddings.

\subsection{Stabilizing policy learning}

Since the policy gradients from the policy network are not passed through the encoders that are trained in parallel, it is essential for the encoders to learn stable embeddings throughout their training to facilitate stable training of the policy network. Since the CL objective alone is invariant under invertible transformations of the entire embedding space \cite{vonkugelgen2021contentstyle}, we add a stability loss to the objective to avoid large deviations of the latent embeddings during training:
\begin{equation}
    \mathcal L_{\text{stability}} = 
    -\frac{1}{N} \sum_{i=1}^{N} 
    \left(
    \mathrm{sim}(l_{T, \text{old}}^i, l_T^i) + \mathrm{sim}(l_{S, \text{old}}^i, l_S^i)
    \right)
\end{equation}
where $l_T^i$ and $l_S^i$ are the raw logits of the policy network while $l_{T, \text{old}}$ and $l_{S, \text{old}}$ are the same before the training phase started. Although this term is computed by processing the observations through both encoder and policy networks, it serves to stabilize the embedding space and thus its gradients are only used to update the encoder networks. This specific formulation encourages updates in the encoder network that will still produce similar actions under the policy network while giving enough flexibility to the embedding space to reorient itself for newly encountered observation samples.

With these terms combined, the encoder networks are trained using gradient descent to minimize the following objective:
\begin{equation}
    \mathcal L_{\text{embedding}} = \mathcal L_\text{contrastive} + \mathcal L_\text{alignment} + \mathcal L_\text{stability}
\end{equation}
This term is optimized using aligned trajectory dataset collected by rolling out the teacher policy, which is trained in parallel with RL to maximize the task objective.

\section{Experiments}
\label{sec:experiments}

\begin{figure}
\centering
\includegraphics[scale=0.66]{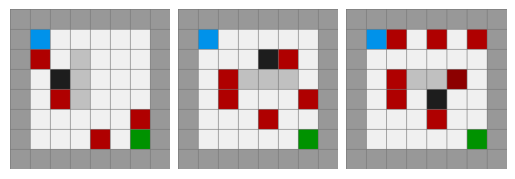}
\caption{TunnelVision environment with the agent (black), start (cyan) and goal (green) coordinates with randomly generated obstacle configurations (red). Teacher can observe all 8 surrounding cells while the student sees only 3 adjacent cells in front of it (highlighted in gray), direction of which is controlled by the agent's orientation.}
\label{fig:tunnelvision}
\end{figure}

\begin{figure}
\centering
\includegraphics[scale=0.5]{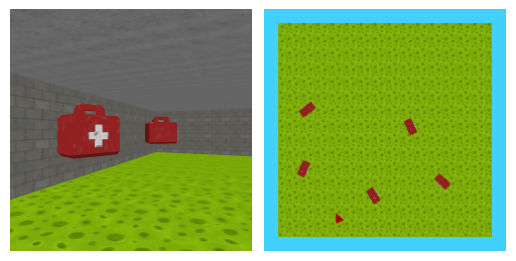}
\caption{CollectHealth environment from (a) the student's perspective and (b) in bird-eye view. Teacher can observe the exact box locations but needs to reorient itself towards the next target to enable the student to imitate it.}
\label{fig:collecthealth}
\end{figure}

\subsection{Environments}
We conduct experiments in two different simulation environments that highlight the imitation gap between policies. In both of the environments, the optimal policy for the teacher agent trained in isolation cannot be imitated by the student due to its relatively limited information about the environment. 

\textbf{CollectHealth \cite{towers2024gymnasium}} The agent needs to collect five health boxes that are randomly spawned in a 3D $15\text{m} \times 15\text{m}$ room (see Fig. \ref{fig:collecthealth}). The agent can move in eight discrete directions or change the camera's orientation by $\pm15 ^\circ$ ($(\mid\mathcal A\mid=10)$). The agent is rewarded for each box it collects and punished by a small amount for every intermediary step. The episode is terminated if all the boxes are collected or the agent hits one of the walls. The teacher agent has access to the exact location of the boxes, which allows it to solve the task without changing its orientation. The student agent observes the environment through limited-FOV FPV images and needs to constantly reorient itself to attend to the next target.

\begin{figure*}
\centering
\includegraphics[scale=0.5]{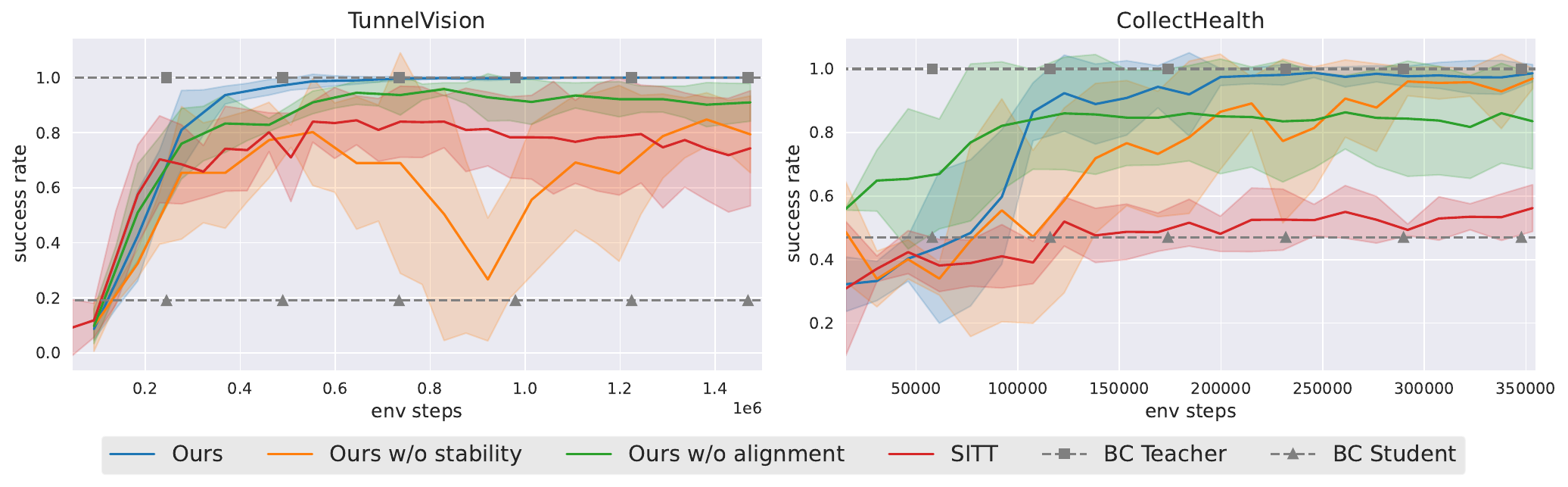}
\caption{Success rate curves for the distilled student policies and state-based teacher policy (BC Teacher) in each environment. Behavior cloning is trained with supervised learning on data collected by a fixed teacher, hence its performance is shown with a horizontal line.}
\label{fig:curves}
\vspace{-0.2cm}
\end{figure*}

\textbf{TunnelVision} This is a grid world environment where the teacher needs to deliberately change its behavior at each timestep to enable the student to imitate it. The agent starting from a fixed position needs to navigate to a fixed goal position while avoiding obstacles that are randomly generated at every episode (see Fig. \ref{fig:tunnelvision}). The agent can independently move or change its orientations in four directions $(\mid\mathcal A\mid=8)$. The agent is rewarded for finishing the task and punished for collisions, upon which the episode is terminated. A small step penalty is added to motivate exploration. The teacher agent observes all eight surrounding cells while the student can only observe three cells in the direction of its orientation.  The time-optimal solution for the teacher is to directly move to the goal without changing its orientation, while the student needs to reorient itself frequently to avoid moving into unobservable obstacles.

\subsection{Baselines}
We compare our method against two competitive imitation learning baselines and do ablation studies to evaluate the effectiveness of our alignment and stability losses: 

\textbf{(1) Behavior Cloning (BC)} We train the teacher policy separately to maximize its own objective, then use supervised action discrepancy loss on the trajectory dataset collected by the teacher to train the student policy:
\[
\mathcal{L}_{\mathrm{BC}} = -\frac{1}{N}\sum^N_{(z_T, z_S) \sim \mathcal{D_T}} 
 \left\| z_T - z_S \right\|^2
\]

\textbf{(2) SITT \cite{messikommer2025student}}  
We jointly train teacher and student and penalize the teacher for visiting states where the student cannot imitate its actions. The exact penalty is computed by a KL divergence between the action distributions and is applied both on the reward signal and the policy gradients:
\begin{equation}
\begin{split}
\tilde{J}(\pi_T) &= 
\mathbb{E}_{s \sim d^{\pi_T},\, a \sim \pi_T(\cdot|s)}\big[r(s,a)\big] \\
&\quad - \alpha \mathbb{E}_{s \sim d^{\pi_T}}\big[
D_{\mathrm{KL}}\big(\pi_T(\cdot|s)\,\|\,\pi_S(\cdot|s)\big)
\big],
\end{split}
\end{equation}
where $\alpha$ is a hyperparameter controlling the tradeoff between task performance and imitation loss. We experiment with various $\alpha$ values and choose the one that yields the highest performance. We use the original implementation of the algorithm.

For a fair comparison, all methods are trained using identical network architectures and the same number of environment interactions. Each experiment is repeated with at least five random seeds, and we report the average success rate. The results are summarized in Table \ref{tab: comparison}. The success rate of the student policies over the training period is shown in Fig.~\ref{fig:curves}.

As expected, the teacher policy trained in isolation exploits private information available to it and hence provides weak imitation signals to the BC-based student policy, which results in high imitation gap. SITT improves upon BC by explicitly penalizing the teacher for visiting states where the student cannot match its behavior, which in turn reduces the imitation gap. However, this comes at the cost of degraded teacher performance, as the modified reward introduces a trade-off between task optimality and imitability.

\begin{table}[h]
\centering
\caption{Success rates of the teacher and student policies in different environments and the imitation gap between them.}
\begin{tabular}{l|ccc|ccc}
\hline
 & \multicolumn{3}{c|}{TunnelVision} & \multicolumn{3}{c}{CollectHealth} \\
Method & T & \textbf{S} & $\Delta$ & T & \textbf{S} & $\Delta$ \\
\hline
BC & 1.00 & 0.19 & 0.81 & 1.00 & 0.47 & 0.53 \\
SITT & 0.91 & 0.72 & 0.19 & 0.94 & 0.51 & 0.43  \\
Ours w/o alignment & 0.99 & 0.91 & 0.08 & 0.94 & 0.73 & 0.21  \\
Ours w/o stability & 0.79 & 0.78 & 0.01 & 0.97 & 0.95 & 0.02  \\
Ours & 1.00 & \textbf{1.00} & \textbf{0.00} & 0.99 & \textbf{0.98} & \textbf{0.01}  \\
\hline
\end{tabular}
\label{tab: comparison}
\end{table}

In contrast, our method enforces imitability at the representation level by restricting both policies to operate on a shared latent space that removes teacher-specific private information. As a result, the teacher is naturally constrained to learn behaviors that are reproducible by the student, which leads to consistently high student performance and a substantially reduced imitation gap without sacrificing the original task objective.

We also observe that removing the alignment loss leads to a noticeable degradation in student performance, especially in CollectHealth, which indicates that explicitly enforcing similarity between paired embeddings is important for effective imitation. Removing the stability loss causes a substantial drop in TunnelVision and a smaller degradation in CollectHealth. This suggests that the stability objective is particularly important in settings with higher sensitivity to representation shifts, where a single mistake can easily fail the episode, while still providing a modest benefit in CollectHealth. Overall, the full model achieves the most consistent performance across environments while maintaining high teacher performance and a minimal imitation gap.

\section{Future work and conclusions}

In this work, we proposed a task-agnostic approach to reduce the imitation gap in RL-driven imitation learning by learning a shared latent representation between teacher and student observation spaces. Our method combines contrastive learning with reinforcement learning to ensure that the teacher policy relies only on the information that is also accessible to the student, which in turn enables the student to imitate it. We also introduce stability and alignment objectives to the embedding space to facilitate steady policy training and tighter imitation gap. Unlike previous approaches, our method is task-agnostic and does not require modifications to the reward function.

Experimental results demonstrate that our approach consistently outperforms baselines and achieves high student performance with minimal imitation gap. Our ablation studies further highlight the effectiveness of our alignment and stability objectives in improving the performance of the distilled student policy. Future work includes extending the method to environments with more complex and high-dimensional continuous control tasks.

\section*{Acknowledgements}
This work was partially supported by the Wallenberg AI, Autonomous Systems and Software Program (WASP) funded by the Knut and Alice Wallenberg Foundation.

\bibliographystyle{IEEEtran}
\bibliography{references}
\end{document}